\documentclass{article}

 \usepackage[preprint]{neurips_2026}

\usepackage[utf8]{inputenc} 
\usepackage[T1]{fontenc}    
\usepackage{hyperref}       
\usepackage{url}            
\usepackage{booktabs}       
\usepackage{amsfonts}       
\usepackage{nicefrac}       
\usepackage{microtype}      
\usepackage{xcolor}         
\usepackage{amssymb} 
\usepackage{wrapfig}
\usepackage{caption}
\usepackage{algorithm}
\usepackage{algpseudocode}
\usepackage{amsmath}
\usepackage{graphicx}
\usepackage{enumitem}
\usepackage{multirow}

\usepackage[table]{xcolor}
\usepackage{amsthm}

\usepackage{soul}
\definecolor{lightyellow}{RGB}{255,248,200}
\sethlcolor{lightyellow}
\definecolor{darkgreen}{RGB}{0,80,60}
\definecolor{darkrose}{RGB}{90,0,120}
\definecolor{darkblue}{RGB}{0,60,100}
\definecolor{darkorange}{RGB}{170,80,20}

\usepackage{mdframed}

\title{Stochastic Meta-Unlearning: Bridging Language Backbone and Multimodal Unlearning}

\author{%
  Zijie Liu$^{1}$, Jinhao Duan$^{1}$, Gaowen Liu$^{2}$, Sijia Liu$^{3}$, Tianlong Chen$^{1}$ \\
  $^{1}$ UNC at Chapel Hill, 
  $^{2}$ Cisco Research,
  $^{3}$ Michigan State University \\
  \texttt{tianlong@cs.unc.edu} \\
}

\begin{document}

\maketitle

\begin{abstract}
Machine unlearning for vision-language models (VLMs) remains underexplored. Unlike language models, VLMs combine a language backbone with visual components, which makes unlearning more complex. There is a surprising phenomenon when moving from single-modality unlearning to VLM unlearning: a target forgotten by the standalone language backbone can still be recovered when image information is given to the full VLM. This shows that text-only feedback is not enough for reliable VLM unlearning.
Motivated by this observation, we propose Stochastic Meta-Unlearning (SMU), a bilevel framework that uses VLM-level feedback to learn an unlearning-ready initialization. In the inner loop, SMU applies a few unlearning steps to the language backbone using text data. In the outer loop, SMU recomposes the updated backbone with the frozen VLM and evaluates forgetting and utility at the VLM level. This design makes the unlearning update aware of the final multimodal behavior, while still keeping the update local to the language backbone.
Experiments on two VLMs, two multimodal meme datasets, and three baselines show that SMU achieves the best overall forget-retain trade-off. Compared with the strongest baseline for each metric, SMU reduces average Forget accuracy by 10.52 points and improves average Retain and Test accuracy by 20.10 and 17.01 points, respectively. More importantly, SMU also transfers to new forgetting targets and to different meta-test unlearning methods. These results suggest that VLM-level feedback can make language-backbone unlearning more reliable and more transferable for VLMs.
\end{abstract}

\section{Introduction}
\label{sec:introduction}
Machine unlearning for vision-language models (VLMs) is both practically important and underexplored. VLMs are increasingly deployed in sensitive settings, such as privacy-sensitive assistance, safety-critical applications, and personalized services, where the ability to selectively remove specific knowledge or behaviors after training is essential~\citep{radford2021learning,liu2023visual,wang2025specify,li2025sauce}. Yet unlearning in VLMs is substantially more complex than in standalone language models, because a VLM couples a visual encoder, a projection module, and a language backbone into a tightly interdependent system
\citep{alayrac2022flamingo,li2023blip,liu2023visual,zhang2024llava}.
Edits to any single component can propagate across modalities in unpredictable ways, making targeted forgetting far harder to achieve.

Existing work addresses this challenge by optimizing the composed VLM directly. This route provides full multimodal supervision and has been applied to visual concept removal, privacy protection, safety alignment, sequential deletion, and modality-aware unlearning \citep{li2024single,dontsov2025clear,liu2025protecting,ma2024benchmarking,kawakami2025pulse,huo2025mmunlearner,liu2025modality,chen2025safeeraser}. However, because the visual encoder, projector, and language backbone must remain aligned, updates targeting one behavior can inadvertently degrade unrelated capabilities, a side effect we confirm experimentally.

\begin{wrapfigure}{r}{0.63\textwidth}
    \centering
    \vspace{-3mm}
    \includegraphics[width=0.62\textwidth]{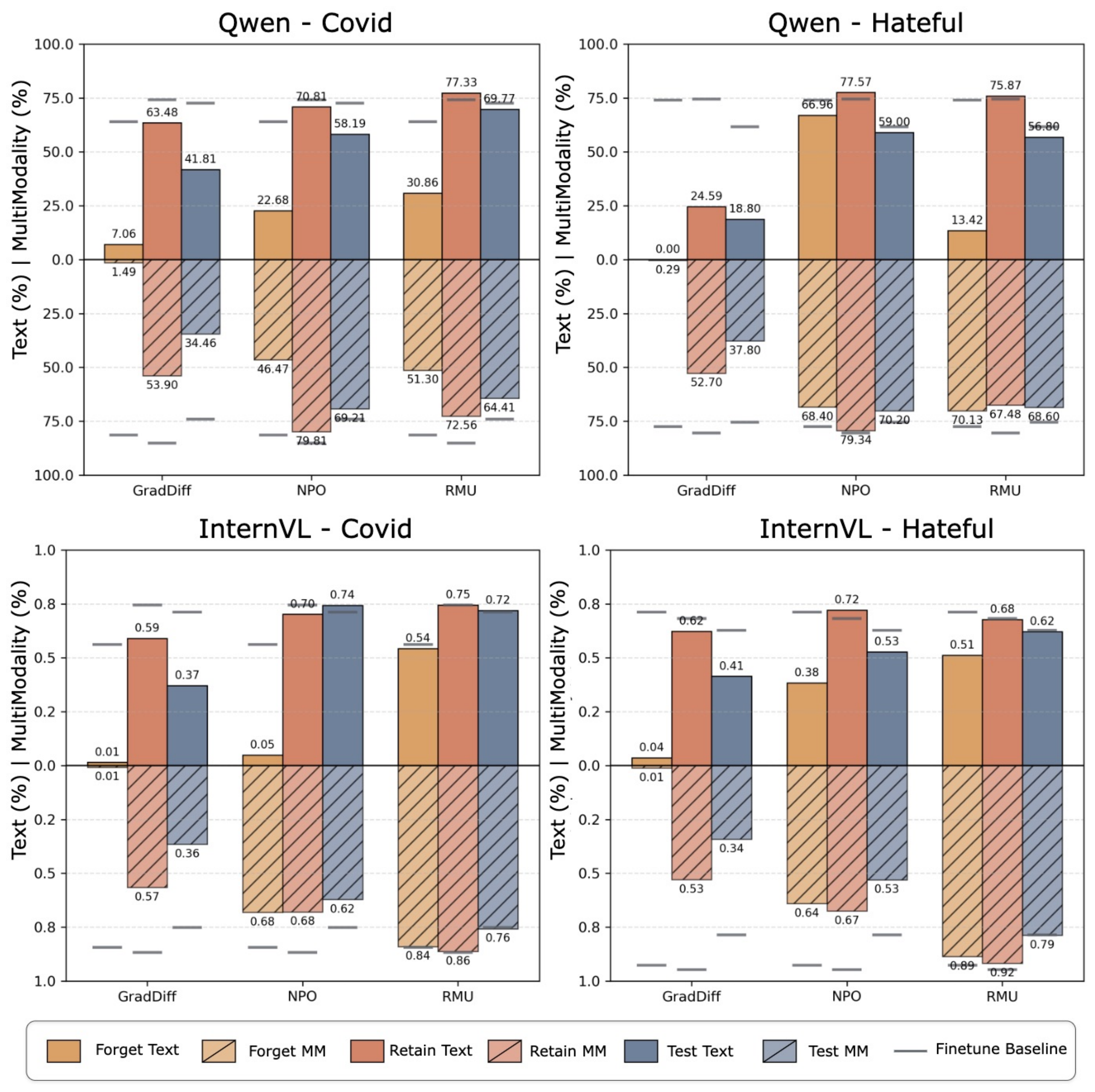}
    \caption{\small \textbf{Text-side unlearning alone does not guarantee multimodal forgetting.}
Across Qwen and InternVL on Covid and Hateful Memes, LM-side unlearning can reduce Forget Text accuracy, but the target often remains recoverable under image-conditioned VLM queries.}
    \label{fig:intro}
    \vspace{-2mm}
\end{wrapfigure}

A natural alternative is to unlearn only the language backbone and recompose it with the frozen visual modules. This approach is appealing: it updates only the language backbone, keeps the visual encoder and projector intact, and can directly reuse existing language-model unlearning objectives \citep{petroni2019language,jiang2020can,wallat2020bertnesia}. However, unlearning is optimized and verified on the standalone language model, with no guarantee that forgetting holds after the backbone is recomposed with visual modules. Prior work shows that behavior suppressed in one modality can remain accessible through another \citep{huo2025mmunlearner,wang2025umu}, and we observe the same failure in Fig.~\ref{fig:intro}: a backbone that forgets the target under text-only queries can recover it when image information is provided at VLM inference time. To our knowledge, no prior work has successfully made backbone-side unlearning hold under multimodal recomposition. 

These observations expose a fundamental mismatch. For utility preservation, the intervention should be local to the language backbone. For correctness, forgetting must be verified in the fully composed VLM. A single-level objective cannot capture both requirements: if defined only on the backbone, the update is blind to multimodal recovery; if defined on the full VLM, the update risks disturbing multimodal utility.

\begin{mdframed}[userdefinedwidth=.99\linewidth,align=center,skipabove=3pt,skipbelow=3pt,innerleftmargin=6pt,innerbottommargin=6pt,innertopmargin=6pt,roundcorner=3pt,backgroundcolor=cyan!5,linecolor=gray]  
\noindent \textbf{Key Challenge.} \textit{How can we preserve the utility of language-backbone unlearning while ensuring forgetting holds in the fully composed VLM?}
\end{mdframed}

To address this challenge, we formulate VLM unlearning as a bilevel meta-optimization problem. The \textbf{inner loop} performs the unlearning operation used at test time: it applies $K$ steps of language-backbone unlearning on unimodal text queries. The \textbf{outer loop} reinserts the updated backbone into the frozen VLM and evaluates forget and retain losses on both text-only and image-conditioned queries. The meta-gradient is backpropagated through the inner unlearning steps, so the visual modules are never updated but are used to assess whether backbone-level forgetting persists after multimodal recomposition.

Based on this formulation, we propose \textbf{Stochastic Meta-Unlearning} (SMU). SMU meta-learns a language backbone initialization such that, after $K$ steps of unimodal unlearning from that initialization, the recomposed VLM forgets the target concept under both text-only and image-conditioned queries while retaining performance on unrelated tasks. To improve generalization across diverse forgetting targets, SMU stochastically samples forgetting tasks during meta-training, exposing the initialization to a distribution of unlearning scenarios rather than a fixed target. At meta-test time, SMU applies only backbone-side unlearning, but the initialization has been optimized with VLM-level multimodal feedback.

We evaluate SMU on two VLMs and two multimodal meme benchmarks, comparing against three direct VLM unlearning baselines under both single-modality and multimodal evaluation. SMU achieves the strongest overall forget-retain trade-off: compared with the strongest baseline per metric, SMU reduces average Forget accuracy by 10.52 points while improving average Retain and Test accuracy by 20.10 and 17.01 points, respectively. We further show that SMU generalizes to unseen forgetting targets and remains robust when the meta-test unlearning method changes.
Our contributions are summarized as follows:
\begin{itemize}
\item We identify a component-system mismatch in VLM unlearning and propose \textbf{Stochastic Meta-Unlearning} (SMU), a bilevel framework that addresses it: the inner loop performs language-backbone unlearning on unimodal queries, while the outer loop uses the recomposed VLM to provide multimodal forget-retain feedback, ensuring that backbone-level forgetting holds after recomposition.
\item We show that the stochastic task sampling in SMU addresses the generalization problem in VLM unlearning: by exposing the backbone initialization to a distribution of forgetting tasks during meta-training, SMU transfers to unseen forgetting targets rather than overfitting to a fixed unlearning scenario.
\item We empirically show that SMU improves the forget-retain trade-off over direct VLM unlearning baselines, generalizes to new forgetting targets, and remains robust when the meta-test unlearning method changes.
\end{itemize}

\section{Related Works}
\textbf{Machine Unlearning in Classical Models and LLMs.}

Machine unlearning was first formulated as removing the influence of selected training data without retraining from scratch, and later developed into practical approximate deletion schemes \citep{cao2015towards,bourtoule2021machine,neel2021descent}. Empirical work on deep unlearning established that forget quality, utility preservation, and repeated-request robustness are distinct desiderata that must be evaluated separately \citep{kurmanji2023towards}. In language models, subsequent studies addressed post-hoc removal of private, copyrighted, or harmful knowledge through privacy-oriented erasure, targeted content removal, and systematic benchmarks \citep{jang2023knowledge,eldan2024s,yao2024large,yao2024machine}. TOFU, RWKU, and MUSE further showed that LLM unlearning must be evaluated beyond exact forget prompts, covering entity knowledge, privacy leakage, adversarial probing, and sequential requests \citep{maini2024tofu,cao2024rwku,shi2024muse}. Optimization-focused methods such as NPO, ULD, ECO, SPUL, and SOUL improve the forget--retain trade-off but still study forgetting within a fixed model and procedure \citep{zhang2024negative,ji2024reversing,liu2024large,bhaila2025soft,jia2024soul}. Analyses of brittleness and reversibility motivate treating transfer as a first-class objective rather than assuming that forgetting under one setup generalizes to another \citep{liu2025rethinking,yuan2024closer,doshi2024does,hu2024unlearning}.

\textbf{Multimodal and Vision-Language Unlearning.}
Recent work extends unlearning to VLMs through new benchmarks and architecture-specific methods \citep{li2024single,dontsov2025clear,liu2025protecting,ma2024benchmarking}. SIU shows that forgetting a visual concept requires carefully constructed multimodal objectives rather than direct reuse of text-only recipes \citep{li2024single}. CLEAR, MLLMU-Bench, FIUBench, and PULSE show that multimodal unlearning is harder to evaluate because target knowledge is distributed across images, text, and alignment behavior, and robustness under adversarial or sequential settings is often weak 
\citep{dontsov2025clear,liu2025protecting,ma2024benchmarking,kawakami2025pulse}. Recent methods address this by directly updating the multimodal model via geometry-constrained optimization, modality-aware neuron pruning, or safety-oriented decoupling losses \citep{huo2025mmunlearner,liu2025modality,chen2025safeeraser}. Privacy-centered evaluations confirm that forgetting in MLLMs must be checked from both unimodal and multimodal views under stronger attack protocols \citep{patil2025unlearning,liu2025protecting,ma2024benchmarking}. In contrast, our method meta-trains the LM backbone so that a small number of text-side unlearning steps transfers into the desired multimodal forget/retain behavior after composition with a frozen VLM.

\textbf{Transferable Unlearning and Meta-Learning.}
A smaller but directly relevant literature examines whether forgetting can transfer across data domains, task configurations, or downstream deployments \citep{sepahvand2024data,lu2024towards}. Meta-learning provides the algorithmic template for optimizing parameters that become useful after a few gradient updates \citep{finn2017model,nichol2018first}, an idea that has recently entered unlearning through learning-to-unlearn and meta-unlearning formulations that prepare the pre-update model for more effective or relearning-robust forgetting \citep{huang2024learning,gao2025meta}. Related settings such as unlearning personal data from a single image further underscore the need for models prepared for post-hoc forgetting under restricted forget-data access \citep{de2024unlearning}. Our method brings these ideas into the LM-to-VLM setting: the inner loop performs text unlearning on the LM, while the outer loop evaluates multimodal forgetting and retention after the updated LM is plugged into a frozen VLM, learning an unlearning-ready initialization for transferable multimodal forgetting.

\vspace{-2mm}
\section{Problem Setup}
\vspace{-2mm}
\label{sec:pre}

\textbf{Model notation.}
Let $f_{\theta_T}$ denote the language backbone of a VLM, where $\theta_T$ denotes the trainable language-backbone parameters or trainable adapters. Let $\phi$ denote the remaining frozen VLM components, including the visual encoder, projection module, and multimodal input interface. For a multimodal input $x^{\mathrm{mm}}=(v,t)$ with image $v$ and text prompt $t$, the composed VLM is written as $F_{\theta_T,\phi}(x^{\mathrm{mm}})=F_{\theta_T,\phi}(v,t)$. For a text-only prompt $t$, we write $f_{\theta_T}(t)$ for the standalone language-backbone behavior. If the same prompt is evaluated through the VLM interface without an image, we write $F_{\theta_T,\phi}(\varnothing,t)$. This notation separates the updated language component, $\theta_T$, from the frozen VLM components used for multimodal recomposition, $\phi$.

\textbf{Unlearning task.}
An unlearning request specifies a target $z$ to remove while preserving non-target behavior. We write the corresponding task as $\tau_z=(D_{f}^{\mathrm{text},z},D_{r}^{\mathrm{text}},D_{f}^{\mathrm{mm},z},D_{r}^{\mathrm{mm}})$, where $D_f$ contains forget examples associated with target $z$, $D_r$ contains retain examples, and the superscripts indicate text-only or multimodal data. When the target is clear, we omit $z$ for simplicity. The forget sets are used to optimize or evaluate target removal, while the retain sets are used to optimize or evaluate preservation of non-target utility.

We distinguish the support data used for unlearning from the held-out data used for evaluation. Text-side unlearning uses support sets such as $D_{f,\mathrm{sup}}^{\mathrm{text},z}$ and optionally $D_{r,\mathrm{sup}}^{\mathrm{text}}$, while VLM-side unlearning uses $D_{f,\mathrm{sup}}^{\mathrm{mm},z}$ and optionally $D_{r,\mathrm{sup}}^{\mathrm{mm}}$. Evaluation is performed on disjoint held-out sets $D_{\mathrm{eval}}^{\mathrm{text}}=(D_{f,\mathrm{test}}^{\mathrm{text},z},D_{r,\mathrm{test}}^{\mathrm{text}},D_{g,\mathrm{test}}^{\mathrm{text}})$ and $D_{\mathrm{eval}}^{\mathrm{mm}}=(D_{f,\mathrm{test}}^{\mathrm{mm},z},D_{r,\mathrm{test}}^{\mathrm{mm}},D_{g,\mathrm{test}}^{\mathrm{mm}})$, where $D_f$ measures target removal, $D_r$ measures retain utility, and $D_g$ denotes the general held-out test split.

\textbf{Text-side unlearning.}
Given an unlearning method $m\in\mathcal{M}$, text-side unlearning updates only the language backbone. Starting from $\theta_{T,0}=\theta_T$, it applies $K$
updates $\theta_{T,k}=U_{m,\mathrm{text}}(\theta_{T,k-1}; D_{f,\mathrm{sup}}^{\mathrm{text},z},D_{r,\mathrm{sup}}^{\mathrm{text}})$ for $k=1,\ldots,K$, producing $\theta_{T,K}=U_{m,\mathrm{text}}^K(\theta_T; D_{f,\mathrm{sup}}^{\mathrm{text},z},D_{r,\mathrm{sup}}^{\mathrm{text}})$. Some unlearning methods use only forget examples; in that case, $D_{r,\mathrm{sup}}^{\mathrm{text}}$ is omitted. The updated backbone can be evaluated alone as $f_{\theta_{T,K}}$ or reinserted into the frozen VLM and evaluated as $F_{\theta_{T,K},\phi}$. This procedure is computationally simple because the unlearning updates do not require multimodal forward or backward passes, but the objective itself is not optimized through the composed VLM and therefore may not guarantee forgetting under image-conditioned queries.

\textbf{VLM-side unlearning.}
VLM-side unlearning computes the unlearning loss through the composed VLM. In our setting, the non-language VLM components $\phi$ remain fixed, and only the language-backbone parameters $\theta_T$ are updated. Starting from $\theta_{T,0}^{\mathrm{vlm}}=\theta_T$, VLM-side unlearning applies $\theta_{T,k}^{\mathrm{vlm}}=
U_{m,\mathrm{vlm}}(\theta_{T,k-1}^{\mathrm{vlm}};
\phi,D_{f,\mathrm{sup}}^{\mathrm{mm},z},D_{r,\mathrm{sup}}^{\mathrm{mm}})$ for $k=1,\ldots,K$, yielding
$\theta_{T,K}^{\mathrm{vlm}}=
U_{m,\mathrm{vlm}}^K(\theta_T;\phi,
D_{f,\mathrm{sup}}^{\mathrm{mm},z},D_{r,\mathrm{sup}}^{\mathrm{mm}})$. The resulting model is $F_{\theta_{T,K}^{\mathrm{vlm}},\phi}$. The difference from text-side unlearning is not that $\phi$ is updated, but that the unlearning objective is evaluated through $F_{\theta_T,\phi}$ on multimodal inputs, which requires VLM-level computation at each unlearning step.

\textbf{Evaluation settings.}
After any unlearning method produces an updated backbone $\bar{\theta}_T$, we evaluate it in two settings. \textbf{Text$\rightarrow$Text} evaluates the standalone backbone $f_{\bar{\theta}_T}$ on $D_{\mathrm{eval}}^{\mathrm{text}}$, where $D_{f,\mathrm{test}}^{\mathrm{text},z}$ measures whether the target is removed under language-only prompts, $D_{r,\mathrm{test}}^{\mathrm{text}}$ measures retained language utility, and $D_{g,\mathrm{test}}^{\mathrm{text}}$ measures general text-side performance. \textbf{Text$\rightarrow$Multi} evaluates the recomposed VLM $F_{\bar{\theta}_T,\phi}$ on $D_{\mathrm{eval}}^{\mathrm{mm}}$, where $D_{f,\mathrm{test}}^{\mathrm{mm},z}$ measures whether the target remains accessible when image information is provided, $D_{r,\mathrm{test}}^{\mathrm{mm}}$ measures retained vision-language utility, and $D_{g,\mathrm{test}}^{\mathrm{mm}}$ measures general multimodal performance. Lower Forget accuracy indicates stronger removal, while higher Retain and Test accuracy indicates better utility preservation. This distinction motivates the SMU design in Fig.~\ref{fig:teaser}: SMU performs the unlearning update on the language backbone, then recomposes the updated backbone with the frozen VLM components and uses VLM-level forget-retain feedback to meta-update the initialization.

\vspace{-2mm}
\section{Stochastic Meta-Unlearning}
\vspace{-2mm}
\label{sec:method}

\begin{wrapfigure}{l}{0.63\textwidth}
    \centering
    \vspace{-6mm}
    \includegraphics[width=0.62\textwidth]{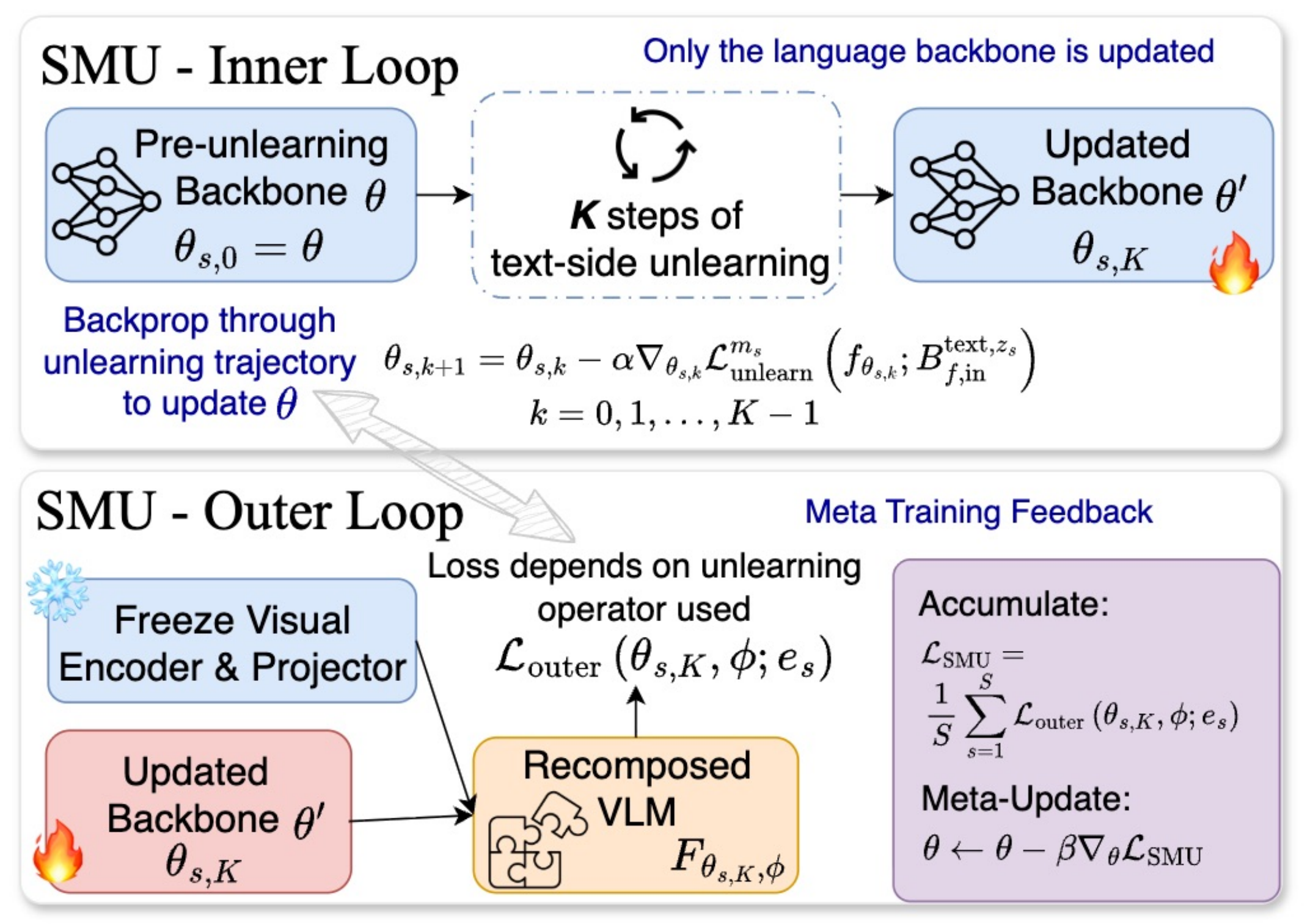}
    \caption{\small \textbf{SMU aligns text-side unlearning with VLM-level forgetting.}
The inner loop updates only the language backbone through text-side unlearning, while the outer loop recomposes the updated backbone with frozen visual components and uses VLM-level feedback to meta-update the initialization.
}
    \label{fig:teaser}
    \vspace{-5mm}
\end{wrapfigure}

We propose \textbf{Stochastic Meta-Unlearning} (SMU), a bilevel framework that aligns language-backbone unlearning with VLM-level forgetting(Figure \ref{fig:teaser}. The inner loop performs the actual unlearning operation on the language backbone using only text-side data. The outer loop then recomposes the updated backbone with the frozen VLM components and evaluates the resulting VLM on multimodal forget and retain queries. The outer loss is backpropagated through the inner unlearning steps to update the pre-unlearning backbone initialization.
Throughout this section, $\theta_T$ denotes the trainable parameters of the language backbone, or the trainable adapters attached to it. We use $\phi$ to denote the frozen non-language VLM components, including the visual encoder, projection module, and multimodal input interface. The standalone language backbone is written as $f_{\theta_T}$, and the recomposed VLM is written as $F_{\theta_T,\phi}$. During inner-loop unlearning, $\theta_{T,k}$ denotes the backbone parameters after the $k$-th unlearning step, and $\theta_T^\star$ denotes the meta-learned initialization returned by SMU.
\vspace{-2mm}
\subsection{Meta-Unlearning Setup}
\vspace{-2mm}
\label{subsec:meta_setup}
Each meta-training episode samples an environment
$e=(z,m,c)\sim p(\mathcal{E})$, where $z$ is the forgetting target, $m$ is the
inner-loop unlearning operator, and $c$ denotes the query condition. For each environment, SMU separates the
data into an inner support batch and outer query batches. The inner support batch
contains text-side examples for applying the unlearning update:
\(B_{\mathrm{in}}^{\mathrm{text},z}=(B_{f,\mathrm{in}}^{\mathrm{text},z},B_{r,\mathrm{in}}^{\mathrm{text}})\),
where the retain batch is optional and is omitted for unlearning operators that
only use forget examples. The outer query batches are used only for
meta-optimization and evaluation after recomposition. They include a multimodal
forget batch $B_f^{\mathrm{mm},z,c}$, a multimodal retain batch
$B_r^{\mathrm{mm},c}$, and optionally text query batches
$B_{f,\mathrm{out}}^{\mathrm{text},z}$ and $B_{r,\mathrm{out}}^{\mathrm{text}}$.
The goal is to learn an initialization $\theta_T$ such that a small number of
text-side unlearning steps produces an updated backbone that forgets the target after VLM recomposition while preserving non-target utility.

\subsection{Meta-Unlearning}
\vspace{-2mm}
\label{subsec:meta_unlearning}

We first describe the base meta-unlearning objective for a fixed environment
$e=(z,m,c)$, where $z$ is the forgetting target, $m$ is the
language-backbone unlearning operator, and $c$ denotes the query condition used
for outer-loop evaluation.

\textbf{Inner update.}
Given an environment $e=(z,m,c)$, the inner loop starts from
$\theta_{T,0}=\theta_T$ and applies $K$ language-backbone unlearning steps:
\begin{equation}
    \theta_{T,k+1}
    =
    \theta_{T,k}
    -
    \alpha
    \nabla_{\theta_{T,k}}
    \mathcal{L}^{m}_{\mathrm{in}}
    \left(
        f_{\theta_{T,k}};
        B_{f,\mathrm{in}}^{\mathrm{text},z},
        B_{r,\mathrm{in}}^{\mathrm{text}}
    \right),
    \quad k=0,\ldots,K-1 .
\label{eq:inner_update}
\end{equation}
Here, $\alpha$ is the inner-loop learning rate, and
$\mathcal{L}^{m}_{\mathrm{in}}$ is the text-side unlearning loss induced by
operator $m$. Some operators use both forget and retain examples, while others
use only the forget batch; in the latter case,
$B_{r,\mathrm{in}}^{\mathrm{text}}$ is omitted. We absorb the sign convention
for forgetting into $\mathcal{L}^{m}_{\mathrm{in}}$, so every inner step is
written as gradient descent. We denote the result of the $K$ inner steps by
\[
\theta_{T,K}
=
U_{m,\mathrm{text}}^K
\left(
    \theta_T;
    B_{f,\mathrm{in}}^{\mathrm{text},z},
    B_{r,\mathrm{in}}^{\mathrm{text}}
\right).
\]
The inner update uses only text-side data and updates only the trainable
language-backbone parameters. It does not require image inputs or
backpropagation through the visual encoder or projection module.

\textbf{Outer feedback.}
After the inner update, SMU recomposes the updated backbone with the frozen VLM
components and evaluates $F_{\theta_{T,K},\phi}$. For each environment $e$, the
outer objective is
\begin{equation}
\begin{aligned}
    \mathcal{L}_{\mathrm{outer}}(\theta_T;e)
    =
    &\lambda_f^{\mathrm{mm}}
    \mathcal{L}_f^{\mathrm{mm}}
    \left(
        F_{\theta_{T,K},\phi};
        B_f^{\mathrm{mm},z,c}
    \right)
    +
    \lambda_r^{\mathrm{mm}}
    \mathcal{L}_r^{\mathrm{mm}}
    \left(
        F_{\theta_{T,K},\phi};
        B_r^{\mathrm{mm},c}
    \right) \\
    &+
    \lambda_f^{\mathrm{text}}
    \mathcal{L}_f^{\mathrm{text}}
    \left(
        f_{\theta_{T,K}};
        B_{f,\mathrm{out}}^{\mathrm{text},z}
    \right)
    +
    \lambda_r^{\mathrm{text}}
    \mathcal{L}_r^{\mathrm{text}}
    \left(
        f_{\theta_{T,K}};
        B_{r,\mathrm{out}}^{\mathrm{text}}
    \right).
\end{aligned}
\label{eq:outer_objective}
\end{equation}
The forget losses encourage removal of the target behavior, and the retain
losses preserve non-target utility. If text-side outer supervision is not used,
we set $\lambda_f^{\mathrm{text}}=\lambda_r^{\mathrm{text}}=0$. The multimodal
terms are the key part of the meta-unlearning objective: although the inner
update is performed on the language backbone, the outer loss evaluates the
updated backbone inside the recomposed VLM.

\textbf{Meta-update.}
The meta-update is
\begin{equation}
    \theta_T
    \leftarrow
    \theta_T
    -
    \beta
    \frac{d}{d\theta_T}
    \mathcal{L}_{\mathrm{outer}}(\theta_T;e),
\label{eq:meta_update}
\end{equation}
where $\beta$ is the meta-learning rate. This derivative is a meta-gradient
because $\theta_{T,K}$ depends on $\theta_T$ through the $K$ inner unlearning
steps. During this update, gradients may flow through the frozen VLM computation
graph to update $\theta_T$, but $\phi$ itself is never updated. In this way,
meta-unlearning learns an initialization whose backbone-side unlearning steps
are judged by VLM-level feedback.

\subsection{Meta-Test Unlearning}
\label{subsec:post_initialization_unlearning}

After meta-training, SMU returns an unlearning-ready initialization
$\theta_T^\star$. At meta-test time, given a new deletion request
$z_{\mathrm{test}}$, a text-side support forget set
$D_{f,\mathrm{sup}}^{\mathrm{text},z_{\mathrm{test}}}$, an optional retain
support set $D_{r,\mathrm{sup}}^{\mathrm{text}}$, and a meta-test unlearning
operator $m_{\mathrm{test}}$, we perform only language-backbone unlearning:
\[
    \theta_{T,K_{\mathrm{test}}}^{\mathrm{test}}
    =
    U_{m_{\mathrm{test}},\mathrm{text}}^{K_{\mathrm{test}}}
    \left(
        \theta_T^\star;
        D_{f,\mathrm{sup}}^{\mathrm{text},z_{\mathrm{test}}},
        D_{r,\mathrm{sup}}^{\mathrm{text}}
    \right).
\]
The final deployed model is the recomposed VLM
$F_{\theta_{T,K_{\mathrm{test}}}^{\mathrm{test}},\phi}$. We then evaluate this
model on held-out text-only and multimodal test sets. Here, meta-test refers to
the unlearning stage after meta-training, where the deletion target or the
unlearning operator may differ from those used during meta-training.
Algorithm~\ref{alg:meta_test} in Appendix~\ref{app:algorithms} summarizes this procedure.

\subsection{Stochastic Meta-Unlearning}
\vspace{-2mm}
\label{subsec:stochastic_meta_unlearning}
\begin{wrapfigure}{r}{0.53\textwidth}
    \centering
    \vspace{-3mm}
    \includegraphics[width=0.52\textwidth]{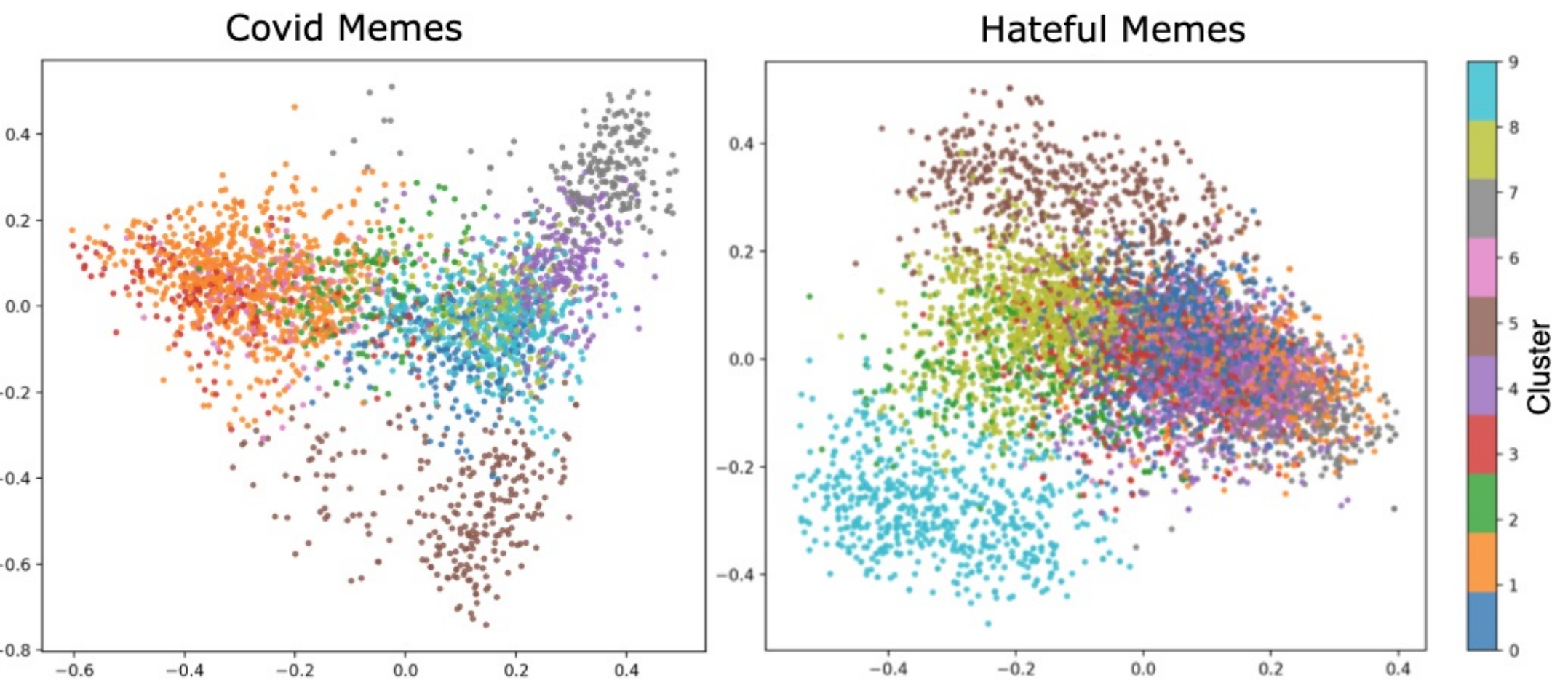}
    \caption{\small \textbf{Cluster-based splits produce coherent forget targets.}
We visualize PCA-projected training examples for Covid Memes and Hateful Memes; colors denote clusters, from which target clusters are selected as forget sets.}
    \label{fig:dataset_pca}
    \vspace{-4mm}
\end{wrapfigure}
The base meta-unlearning objective can overfit to a fixed deletion setting. For example, it may learn an initialization that works for one
forgetting target, one unlearning operator, or one query condition, but does not
transfer when any of them changes. This is undesirable for VLM unlearning,
because real deletion requests may involve new targets, and the unlearning
operator used after initialization may also differ from the one used during
meta-training.

To improve generalization, SMU treats the environment $e=(z,m,c)$ as a random
variable sampled from an environment distribution $p(\mathcal{E})$. The target
$z$ controls what should be forgotten, the operator $m$ controls how the
language backbone is updated in the inner loop, and the condition $c$ controls
how the recomposed VLM is queried in the outer loop. SMU therefore minimizes the
expected outer loss
\(\min_{\theta_T}\mathbb{E}_{e\sim p(\mathcal{E})}
[\mathcal{L}_{\mathrm{outer}}(\theta_T;e)]\).
In practice, each meta-update samples $S$ environments
$\{e_s\}_{s=1}^{S}$ and optimizes the empirical meta-loss
\(\mathcal{L}_{\mathrm{SMU}}^{(S)}(\theta_T)
=
\frac{1}{S}\sum_{s=1}^{S}
\mathcal{L}_{\mathrm{outer}}(\theta_T;e_s)\),
where $e_s\sim p(\mathcal{E})$. The meta-update is
\(\theta_T \leftarrow \theta_T -
\beta \frac{d}{d\theta_T}
\mathcal{L}_{\mathrm{SMU}}^{(S)}(\theta_T)\).
This stochastic design is what enables SMU to target transferability. Sampling
different $z$ encourages transfer to new forgetting targets; sampling different
$m$ encourages robustness to different unlearning operators; and sampling
different $c$ encourages robustness across query conditions. Deterministic
MU is a special case: when $p(\mathcal{E})$ places all
its mass on a single fixed environment, SMU reduces to the fixed-environment
meta-unlearning objective in Sec.~\ref{subsec:meta_unlearning}. Algorithm
\ref{alg:smu} in Appendix~\ref{app:algorithms} summarizes the full stochastic meta-training procedure.

\vspace{-4mm}
\section{Experiments}
\vspace{-2mm}
\label{sec:exp}

In this section, we demonstrate the advantages of SMU from three perspectives:
(1) whether SMU improves the forget-retain trade-off over direct VLM unlearning baselines; (2) whether SMU can generalize when the forget target changes at meta-test time; and (3) whether SMU remains robust when the meta-test unlearning operator changes under a fixed forget target. 
Together, these experiments test whether SMU provides a more effective and robust unlearning strategy for modular VLMs. We include additional image-side unlearning results in Appendix~\ref{app:image_side}.

\textbf{Models, Datasets and Baselines.}
We evaluate our method using Qwen2.5-VL-7B-Instruct (Qwen25) and InternVL3-8B-hf (InternVL3) on two multimodal meme benchmarks: Hateful Memes(Hateful)~\cite{kiela2020hateful} and COVID Memes(COVID)~\cite{cuo2022understanding}. Each example contains both textual and visual information, which allows us to test whether unlearning performed only on text transfers to multimodal evaluation after the language backbone is inserted back into the VLM. We compare against standard unlearning objectives, including GradDiff~\cite{yao2023large},NPO~\cite{zhang2024negative}, and RMU~\cite{li2024wmdp}.
\begin{wrapfigure}{r}{0.60\textwidth}
    \centering
    \vspace{-3mm}
    \includegraphics[width=0.59\textwidth]{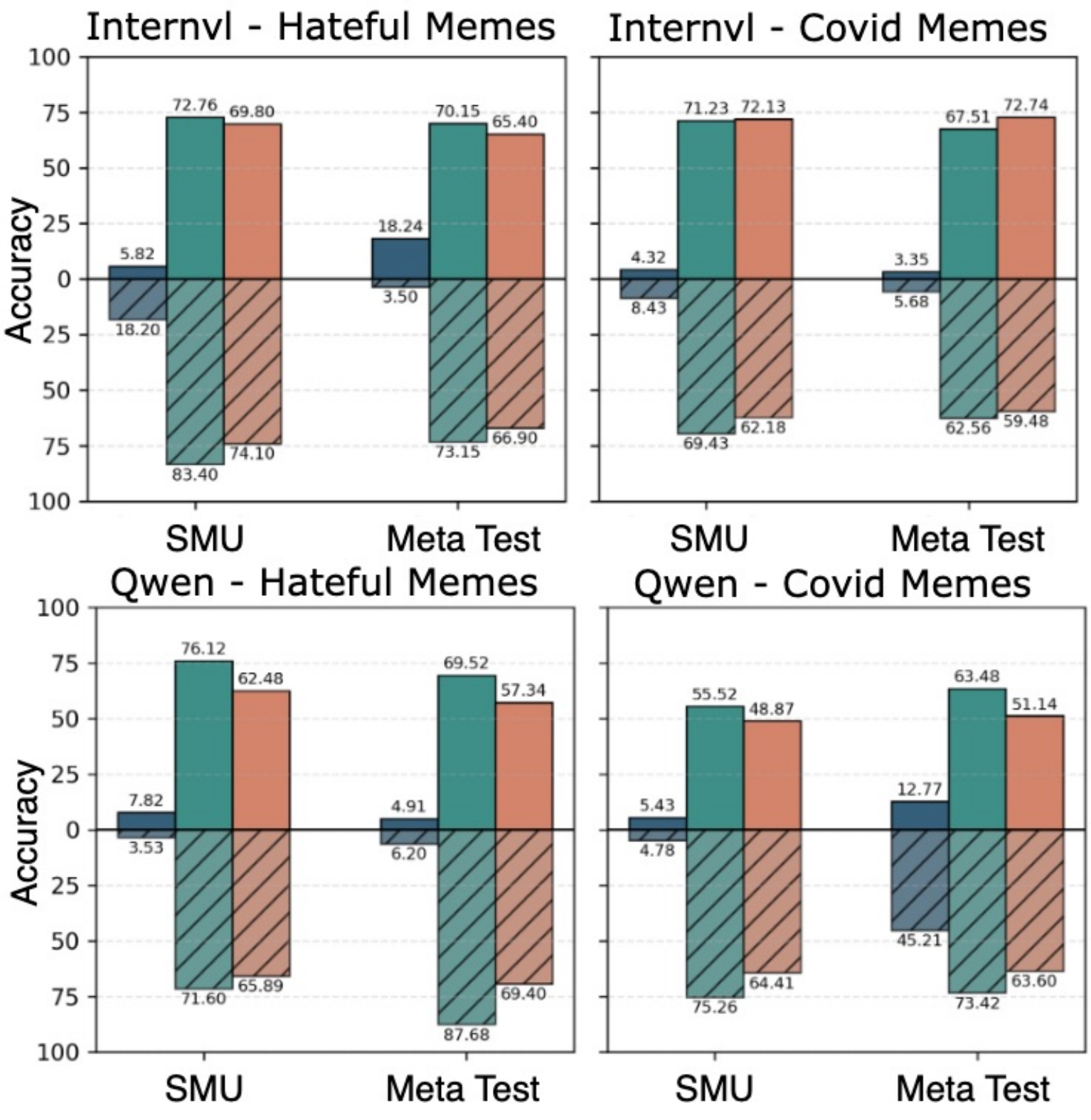}
    \caption{\small \textbf{SMU transfers to new forgetting targets.}
When the meta-test forget target differs from the meta-training target, SMU usually preserves low Forget accuracy while maintaining Retain and Test utility across models and datasets.}
    \label{fig:target_transfer}
    \vspace{-6mm}
\end{wrapfigure}

\textbf{Forget and retain split construction.}
For each dataset, we construct the forget set using a cluster-based procedure. We first represent the training examples in a feature space and apply PCA to the training representations. We then cluster the PCA-projected training examples and select the target cluster as the forget set $\mathcal{D}_{f}$. All remaining training examples are used as the retain set $\mathcal{D}_{r}$. This construction gives a semantically coherent forget set while keeping the retain set disjoint from the examples targeted for removal.

\textbf{Training and unlearning protocol.}
All methods start from the same fine-tuned model. Specifically, we first fine-tune the model on the full training set $\mathcal{D}_{f} \cup \mathcal{D}_{r}$. Unlearning is then applied to this fine-tuned checkpoint. For our method, the inner unlearning update is performed only on text-side data from the language backbone. After unlearning, the updated backbone is plugged back into the frozen VLM, and we evaluate both text-only and multimodal behavior. 

\vspace{-2mm}
\subsection{Main Results}
\vspace{-2mm}
\label{sec:main_results}
\textbf{RQ1: Does SMU outperform direct VLM-side unlearning?}
\begin{table*}[t]
\centering
\caption{
\textbf{Unlearning performance on Hateful Memes and Covid Memes with VLM baselines. Results are accuracy
(\%).} Text $\rightarrow$ Text evaluates text-only forgetting, and Text
$\rightarrow$ Multi evaluates forgetting after VLM recomposition. Lower Forget
is better ($\downarrow$), and higher Retain/Test is better ($\uparrow$). SMU
achieves the strongest overall forget-retain trade-off across models and
datasets.
}
\label{tab:vlm_baseline}
\resizebox{\textwidth}{!}{
\begin{tabular}{l|ccc|ccc||ccc|ccc}
\toprule
\multirow{2}{*}{Method}
& \multicolumn{6}{c||}{Qwen}
& \multicolumn{6}{c}{InternVL} \\
\cmidrule(lr){2-7} \cmidrule(lr){8-13}
& \multicolumn{3}{c|}{Text $\rightarrow$ Text}
& \multicolumn{3}{c||}{Text $\rightarrow$ Multi}
& \multicolumn{3}{c|}{Text $\rightarrow$ Text}
& \multicolumn{3}{c}{Text $\rightarrow$ Multi} \\
& Forget$\downarrow$ & Retain$\uparrow$ & Test$\uparrow$
& Forget$\downarrow$ & Retain$\uparrow$ & Test$\uparrow$
& Forget$\downarrow$ & Retain$\uparrow$ & Test$\uparrow$
& Forget$\downarrow$ & Retain$\uparrow$ & Test$\uparrow$ \\
\midrule

\multicolumn{13}{c}{\textbf{Hateful Memes}} \\
\midrule
\rowcolor{yellow!20}
SMU
& 9.81 & \textbf{78.72} & 68.60
& 20.78 & \textbf{72.32} & \textbf{65.20}
& \textbf{5.82} & \textbf{72.76} & \textbf{69.80}
& \textbf{18.20} & \textbf{83.40} & \textbf{74.10} \\

GradDiff
& \textbf{2.02} & 32.09 & 41.20
& \textbf{0.87} & 31.16 & 40.20
& 46.67 & 61.38 & 59.40
& 64.07 & 60.33 & 67.60 \\

NPO
& 64.07 & 60.19 & 62.20
& 64.21 & 52.13 & 60.00
& 39.41 & 63.01 & 60.60
& 73.30 & 67.68 & 70.00 \\

RMU
& 73.02 & 66.79 & \textbf{69.60}
& 70.13 & 65.08 & 64.20
& 59.99 & 65.38 & 60.60
& 77.06 & 74.52 & 70.20 \\

\midrule

\multicolumn{13}{c}{\textbf{Covid Memes}} \\
\midrule
\rowcolor{yellow!20}
SMU
& 3.35 & \textbf{73.03} & \textbf{66.95}
& 2.60 & \textbf{80.98} & \textbf{68.64}
& \textbf{4.32} & \textbf{71.23} & \textbf{72.13}
& \textbf{8.43} & \textbf{69.43} & \textbf{62.18} \\

GradDiff
& \textbf{0.37} & 8.89 & 5.08
& \textbf{0.00} & 0.00 & 0.00
& 34.94 & 52.66 & 49.72
& 8.55 & 23.00 & 18.64 \\

NPO
& 21.19 & 67.71 & 62.43
& 7.06 & 34.11 & 28.53
& 35.32 & 53.32 & 50.00
& 8.92 & 23.29 & 17.80 \\

RMU
& 21.56 & 64.40 & 56.21
& 2.60 & 19.97 & 13.56
& 32.34 & 57.07 & 52.82
& 10.04 & 27.88 & 22.32 \\

\bottomrule
\end{tabular}
}
\end{table*}
We first ask whether directly optimizing the VLM is the best strategy for unlearning.
Direct VLM-side unlearning is a natural baseline because its loss is computed on the composed multimodal model. However, it does not separate the component being updated from the system being evaluated, so it can remove the target behavior while also damaging retain and test utility.
\begin{wrapfigure}{l}{0.63\textwidth}
    \centering
    \vspace{-3mm}
    \includegraphics[width=0.62\textwidth]{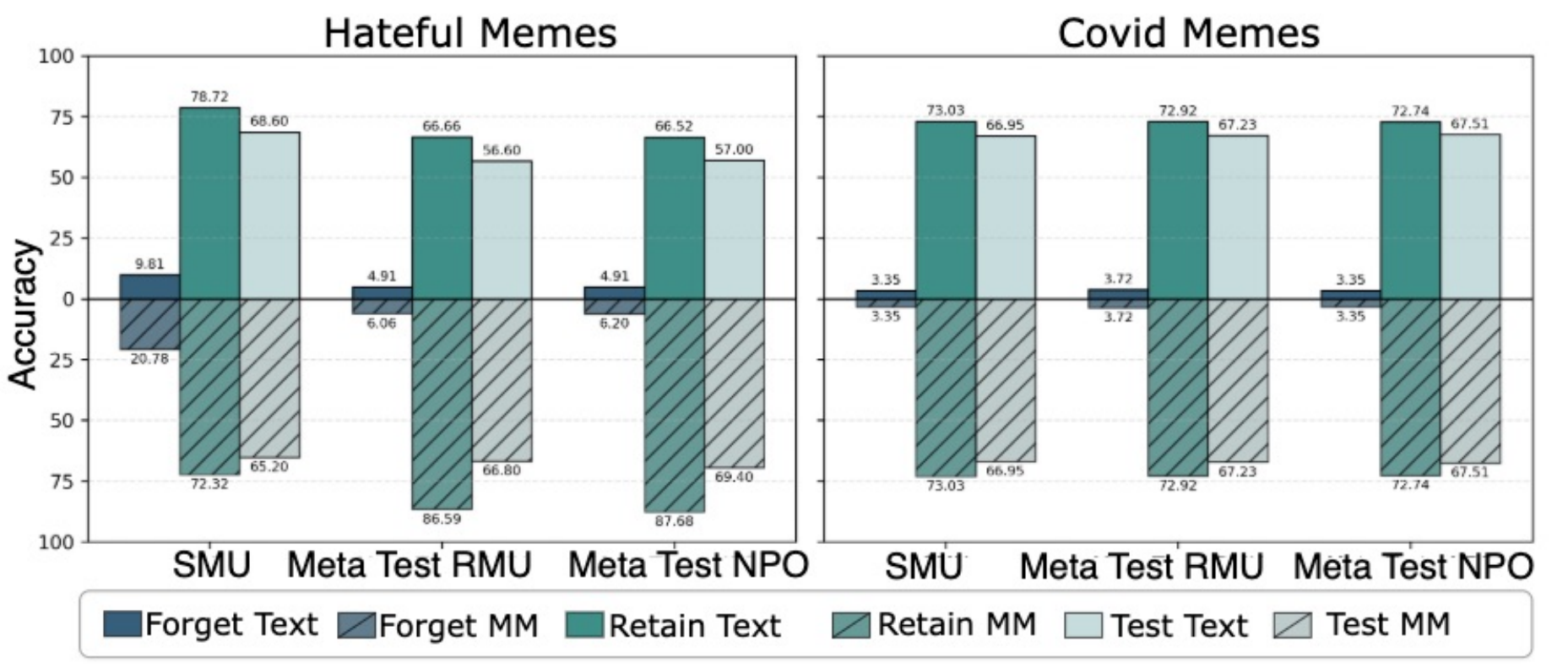}
    \caption{\small \textbf{SMU transfers across meta-test unlearning operators.}
After meta-training with GradDiff, SMU remains effective when the meta-test operator is replaced by RMU or NPO, maintaining low Forget accuracy and high Retain/Test accuracy.}
\label{fig:method_transfer}
    \vspace{-3mm}
\end{wrapfigure}

SMU takes a different approach: it performs the inner unlearning update on the language backbone while using the recomposed VLM for outer-loop forget-retain feedback.
We compare SMU against three direct VLM-side baselines (GradDiff, NPO, RMU) on Hateful Memes and Covid Memes with two backbones (Qwen, InternVL), reporting both text-only (Text $\rightarrow$ Text) and multimodal (Text $\rightarrow$ Multi) evaluation.
Lower Forget accuracy indicates stronger unlearning; higher Retain and Test accuracy indicates better utility preservation.

Table~\ref{tab:vlm_baseline} shows that SMU achieves the best overall forget-retain trade-off.
Averaged over all datasets, models, and evaluation settings, SMU obtains the lowest Forget accuracy ($9.16\%$) while maintaining the highest Retain and Test accuracy ($75.23\%$ and $68.45\%$).
The direct baselines are less balanced. GradDiff sometimes reaches lower Forget accuracy, e.g., $2.02\%$ and $0.87\%$ on Hateful Memes with Qwen, but its Retain/Test accuracy collapses to $32.09\%/41.20\%$ and $31.16\%/40.20\%$; on Covid Memes with Qwen under Text $\rightarrow$ Multi, GradDiff drives all three metrics to $0.00\%$.
NPO and RMU preserve utility better than GradDiff in some settings, but leave substantially higher Forget accuracy: on Hateful Memes with InternVL under Text $\rightarrow$ Multi, both reach $73.30\%$ and $77.06\%$ Forget, while SMU reduces it to $18.20\%$ with the highest Retain/Test accuracy of $83.40\%/74.10\%$.

These results support the bilevel design of SMU.
VLM-level feedback during meta-training aligns the backbone-side unlearning trajectory with final multimodal behavior, enabling SMU to remove target content more consistently while preserving non-target utility across both datasets and backbones.

\textbf{RQ2: Does SMU transfer to new forgetting targets?}
We next study whether SMU can handle a new deletion request at meta-test time.
This setting tests target transferability: the forgetting target used during
meta-training is different from the target used during meta-test. A method that
only works for the meta-training target may overfit to a fixed deletion request,
rather than learning a reusable pre-unlearning initialization.

\begin{wrapfigure}{r}{0.43\textwidth}
    \centering
    \vspace{-3mm}
    \includegraphics[width=0.42\textwidth]{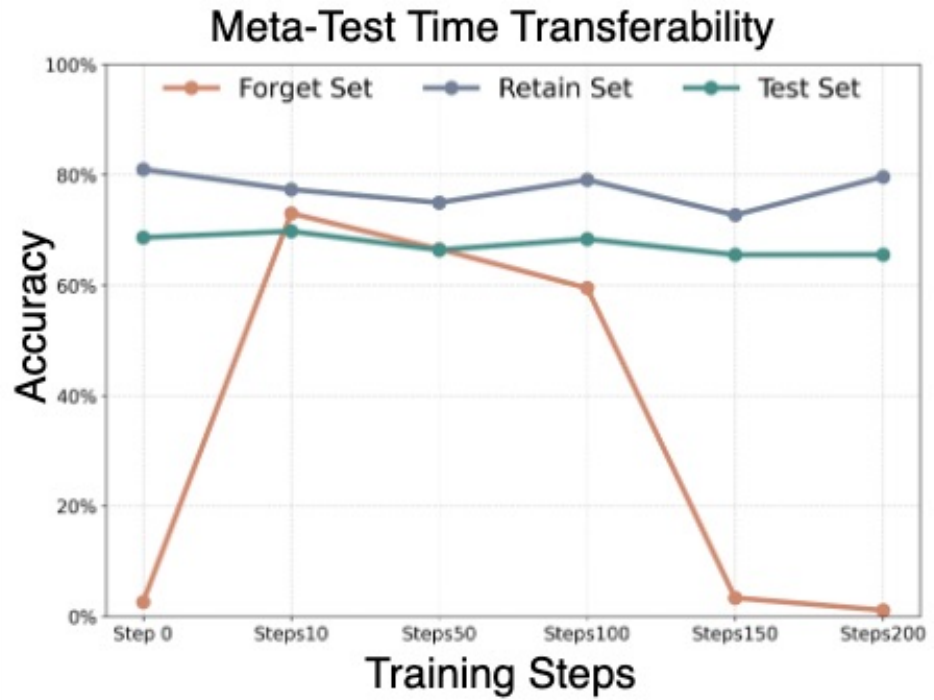}
   \caption{\small \textbf{Operator transfer improves with meta-training.}
Evaluating SMU at different checkpoints shows that Forget accuracy is non-monotonic, while later checkpoints achieve stronger forgetting without large Retain/Test degradation.}
    \label{fig:operator_transfer_steps}
    \vspace{-3mm}
\end{wrapfigure}
To evaluate this, we meta-train SMU on one target and then change the forget
target at meta-test time. Starting from the learned initialization, we perform
the same text-side unlearning procedure on the new target, insert the updated
backbone into the VLM, and evaluate both text-only and multimodal performance.
Lower forget accuracy indicates stronger target removal, while higher retain and test accuracy indicates better utility preservation.
Fig.~\ref{fig:target_transfer} shows that SMU generally transfers to new
forgetting targets. Across most model-dataset settings, the meta-test target
achieves low forget accuracy while retaining useful text and multimodal
performance. For example, on InternVL, SMU keeps forget accuracy low on both
Hateful Memes and Covid Memes after the target changes, while retain and test
accuracy remain stable. On Qwen, SMU also transfers well on Hateful Memes. The
main harder case is Covid under multimodal evaluation, where forget accuracy
increases after target transfer, suggesting that some target shifts are more
difficult when image-conditioned evidence is involved. Overall, the results show
that SMU is not only optimized for one fixed forgetting target. It learns an
initialization that can be adapted to new deletion requests through text-side
unlearning.

\textbf{RQ3: Does SMU transfer across unlearning operators under a fixed target?}
We test whether the learned initialization is tied to the inner-loop operator used during meta-training.
SMU is meta-trained with GradDiff as the inner-loop operator; at meta-test time, we replace it with RMU or NPO while keeping the forget target fixed, then evaluate both text-only and multimodal performance.
Fig.~\ref{fig:method_transfer} shows that SMU remains effective under operator substitution.
On both Hateful Memes and Covid Memes, switching to RMU or NPO at meta-test time still yields low Forget Text/MM while Retain and Test accuracy stay high.
The learned initialization therefore acts as a transferable starting point rather than one fitted to a specific inner-loop update rule.
Fig.~\ref{fig:operator_transfer_steps} shows that this transferability emerges gradually.
Forget accuracy is non-monotonic during meta-training, first rising then falling, while retain and test accuracy remain stable throughout.
Reliable operator transfer thus requires sufficient meta-training; it is not a property of the base model initialization.

\vspace{-4mm}
\section{Conclusion}
\vspace{-4mm}

We identify a component-system mismatch in VLM unlearning: the update is applied
to the language backbone, while forgetting must hold after the backbone is
recomposed with the visual components. To address this mismatch, we propose
\textbf{Stochastic Meta-Unlearning} (SMU), a bilevel framework that performs
language-backbone unlearning in the inner loop and uses VLM-level forget-retain
feedback in the outer loop to learn an unlearning-ready initialization.
Experiments show that SMU improves the forget-retain trade-off over direct
VLM-side unlearning baselines, transfers to new forgetting targets, and remains
effective when the meta-test unlearning operator changes.

\bibliographystyle{plain}
\bibliography{references}


\appendix


\section{Algorithms}
\label{app:algorithms}
For completeness, Algorithms~\ref{alg:smu} and~\ref{alg:meta_test} provide
pseudocode for stochastic meta-training and meta-test unlearning, respectively.
They are deferred from the main paper because the corresponding updates are
already specified by Eqs.~\eqref{eq:inner_update}--\eqref{eq:meta_update}.

\section{Additional Results: Image-Side Unlearning}
\label{app:image_side}

\textbf{RQ4: Is image-side unlearning a reliable intervention point?}
Since VLMs are modular, a natural alternative to language-backbone unlearning is to apply the update to the image-side components instead. 
We test this directly and evaluate both image-side and multimodal performance.\begin{wraptable}{r}{0.54\textwidth}
\centering
\caption{\textbf{Image-side unlearning is unstable.} Results are accuracy (\%).
Lower Forget is better ($\downarrow$), and higher Retain/Test is better
($\uparrow$).}
\label{tab:image_transfer}
\resizebox{0.54\textwidth}{!}{
\begin{tabular}{l|ccc||ccc}
\toprule
\multirow{2}{*}{Method}
& \multicolumn{3}{c||}{Image}
& \multicolumn{3}{c}{MM} \\
\cmidrule(lr){2-4} \cmidrule(lr){5-7}
& Forget$\downarrow$ & Retain$\rightarrow$ & Test$\uparrow$
& Forget$\downarrow$ & Retain$\rightarrow$ & Test$\uparrow$ \\
\midrule

\multicolumn{7}{c}{\textbf{Hateful Memes}} \\
\cmidrule(lr){1-7}
Finetuned & 92.93 & 94.83 & 81.80 & 90.62 & 92.61 & 80.80 \\
GradDiff  & 2.02  & 32.09 & 41.20 & 0.87  & 31.16 & 40.20 \\
NPO       & 64.07 & 60.19 & 62.20 & 64.21 & 52.13 & 60.00 \\
RMU       & 73.02 & 66.79 & 69.60 & 70.13 & 65.08 & 64.20 \\

\midrule
\multicolumn{7}{c}{\textbf{Covid Memes}} \\
\cmidrule(lr){1-7}
Finetuned & 84.39 & 84.33 & 76.27 & 81.41 & 85.09 & 74.01 \\
GradDiff  & 0.37  & 8.89  & 5.08  & 0.00  & 0.00  & 0.00 \\
NPO       & 21.19 & 67.71 & 62.43 & 6.69  & 34.55 & 28.25 \\
RMU       & 21.56 & 64.40 & 56.21 & 2.23  & 19.97 & 13.28 \\

\bottomrule
\end{tabular}
}
\end{wraptable}
Table~\ref{tab:image_transfer} shows that image-side unlearning yields an unstable forget-retain trade-off. 
GradDiff can strongly reduce forget accuracy but severely damages retain and test utility, most clearly on Covid Memes, where multimodal retain and test accuracy collapse to $0.00\%$.
NPO and RMU preserve more utility but consistently leave high forget accuracy, especially on Hateful Memes.
The image side is therefore not a reliable standalone intervention point: it either fails to remove the target behavior or does so by destroying general utility.
This instability supports our design choice of placing the unlearning update on the language backbone with VLM-level feedback guiding the forget-retain objective.

\section{Implementation Details}
\label{app:implementation}

\subsection{Unlearning Objectives}
\label{app:losses}

We use three unlearning objectives as baselines and as possible inner-loop
operators in SMU: GradDiff, NPO, and RMU. Let
$\mathrm{CE}_{\theta_T}(D)$ denote the sequence negative log-likelihood on a
batch $D$, and let $\theta_T^{\mathrm{ref}}$ denote a frozen reference
backbone. Unless otherwise stated, each unlearning objective has the form
\[
    \mathcal{L}
    =
    \gamma \mathcal{L}_{f}
    +
    \alpha \mathcal{L}_{r},
\]
where $\mathcal{L}_{f}$ is the forget loss and $\mathcal{L}_{r}$ is the retain
loss.

\paragraph{GradDiff.}
GradDiff performs gradient ascent on the forget set by using
\[
    \mathcal{L}_{f}
    =
    -\mathrm{CE}_{\theta_T}(D_f).
\]
The retain loss is either the standard NLL loss
$\mathcal{L}_{r}=\mathrm{CE}_{\theta_T}(D_r)$ or a KL loss against the frozen
reference model.

\paragraph{NPO.}
NPO uses a one-sided DPO-style loss on forget examples. Let
$\ell_{\theta_T}(x)$ be the per-example sequence NLL under the current model
and $\ell_{\mathrm{ref}}(x)$ be the NLL under the frozen reference model. The
forget loss is
\[
    \mathcal{L}_{f}
    =
    -\frac{2}{\beta_{\mathrm{NPO}}}
    \mathbb{E}_{x_f}
    \left[
        \log \sigma
        \left(
            \beta_{\mathrm{NPO}}
            (\ell_{\theta_T}(x_f)-\ell_{\mathrm{ref}}(x_f))
        \right)
    \right].
\]
The retain loss follows the same NLL or KL choice as GradDiff.

\paragraph{RMU.}
RMU matches hidden activations on forget examples to a random control vector.
Let $h_{\theta_T}(x)$ be the hidden activation from the selected transformer
block, and let $c$ be a normalized random control vector. For activations
$a,b\in\mathbb{R}^{B\times T\times d}$ and token mask $m$, define
\[
    \mathcal{A}(a,b,m)
    =
    \frac{1}{B}
    \sum_{i=1}^{B}
    \frac{1}{\max(1,\sum_t m_{it})}
    \sum_t
    m_{it}
    \left(
        \frac{1}{d}
        \sum_{j=1}^{d}
        (a_{itj}-b_{itj})^2
    \right).
\]
The RMU forget loss is
$\mathcal{L}_{f}=\mathcal{A}(h_{\theta_T}(x_f),c,m_f)$.
For retention, we either use the NLL/KL retain loss above or an activation
matching loss
$\mathcal{L}_{r}=
\mathcal{A}(h_{\theta_T}(x_r),h_{\theta_T^{\mathrm{ref}}}(x_r),m_r)$.

\paragraph{SMU objective.}
In SMU, the inner loop applies one of the above objectives using text-side
support data. After $K$ inner updates, the updated backbone is recomposed with
the frozen VLM components, and the outer loss is computed on multimodal forget
and retain batches:
\[
    \mathcal{L}_{\mathrm{meta}}
    =
    \lambda_f \mathcal{L}_{f}^{\mathrm{mm}}
    +
    \lambda_r \mathcal{L}_{r}^{\mathrm{mm}}.
\]

\subsection{Dataset Construction}
\label{app:data}

Forget and retain sets are constructed by clustering training examples. We
encode meme text using \texttt{sentence-transformers/all-MiniLM-L6-v2}; if this
encoder is unavailable, we use TF-IDF features with unigrams and bigrams. We
then run KMeans with $k=10$ and random seed 42. The forget set is the union of
selected clusters, and the retain set contains the remaining clusters. Some
transfer experiments additionally exclude specific clusters from the retain set
to avoid overlap with held-out target clusters. The same split construction is
used for all methods.

\subsection{Prompting and Evaluation}
\label{app:evaluation}

For Hateful Memes, the model predicts whether a meme is hateful or not hateful.
For Covid Memes, the model predicts one of three labels: not harmful, somewhat
harmful, or very harmful. We evaluate both text-only and multimodal settings.
\begin{wrapfigure}{r}{0.6\columnwidth}
\vspace{-4mm}
\begin{minipage}{0.58\columnwidth}
\begin{algorithm}[H]
\caption{\small Stochastic Meta-Unlearning}
\label{alg:smu}
\footnotesize
\begin{algorithmic}[1]

\Require Language-backbone initialization $\theta_T$
\Require Frozen VLM components $\phi$
\Require Environment distribution $p(\mathcal{E})$
\Require Inner-loop steps $K$
\Require Number of sampled environments $S$
\Require Step sizes $\alpha,\beta$
\Require Loss weights $\lambda$

\While{not converged}
    \State Initialize meta-loss
    $\mathcal{L}_{\mathrm{SMU}} \leftarrow 0$

    \For{$s=1$ to $S$}
        \State Sample environment
        $e_s=(z_s,m_s,c_s)\sim p(\mathcal{E})$

        \State Sample inner text batch
        $(B_{f,\mathrm{in}}^{\mathrm{text},z_s},
        B_{r,\mathrm{in}}^{\mathrm{text}})$

        \State Set $\theta_{T,s,0}\leftarrow \theta_T$

        \For{$k=0$ to $K-1$}
            \State Compute language-backbone unlearning loss
            \[
                \mathcal{L}^{m_s}_{\mathrm{in}}
                \left(
                    f_{\theta_{T,s,k}};
                    B_{f,\mathrm{in}}^{\mathrm{text},z_s},
                    B_{r,\mathrm{in}}^{\mathrm{text}}
                \right)
            \]

            \State Update language backbone
            \[
                \theta_{T,s,k+1}
                =
                \theta_{T,s,k}
                -
                \alpha
                \nabla_{\theta_{T,s,k}}
                \mathcal{L}^{m_s}_{\mathrm{in}}
                \left(
                    f_{\theta_{T,s,k}};
                    B_{f,\mathrm{in}}^{\mathrm{text},z_s},
                    B_{r,\mathrm{in}}^{\mathrm{text}}
                \right)
            \]
        \EndFor

        \State Recompose updated backbone with frozen VLM:
        $F_{\theta_{T,s,K},\phi}$

        \State Sample outer forget and retain batches under condition $c_s$

        \State Compute outer loss
        $\mathcal{L}_{\mathrm{outer}}^{(s)}
        =
        \mathcal{L}_{\mathrm{outer}}(\theta_T;e_s)$
        using Eq.~\eqref{eq:outer_objective}

        \State Accumulate
        \[
            \mathcal{L}_{\mathrm{SMU}}
            \leftarrow
            \mathcal{L}_{\mathrm{SMU}}
            +
            \frac{1}{S}
            \mathcal{L}_{\mathrm{outer}}^{(s)}
        \]
    \EndFor

    \State Meta-update initialization
    \[
        \theta_T
        \leftarrow
        \theta_T
        -
        \beta
        \nabla_{\theta_T}
        \mathcal{L}_{\mathrm{SMU}}
    \]

\EndWhile

\State \Return meta-learned initialization
$\theta_T^\star \leftarrow \theta_T$

\end{algorithmic}
\end{algorithm}
\end{minipage}
\vspace{-8mm}
\end{wrapfigure}
For text-only evaluation, the prompt contains only the meme text. For multimodal
evaluation, the prompt includes both the image and the meme text. We parse
Hateful Memes predictions by first checking for class labels 0 or 1 and then
falling back to string matching for ``not hateful'' and ``hateful''. For Covid
Memes, we use normalized substring matching against the three class names.
Invalid predictions are treated as unparsable outputs. The main paper reports
accuracy; our evaluation code also computes macro precision, macro recall, and
macro F1.

\subsection{Training Hyperparameters}
\label{app:hyperparameters}

All models are fine-tuned before unlearning. Fine-tuning uses LoRA with rank 8,
learning rate $1\times10^{-4}$, cosine scheduling, warmup ratio 0.1, batch size
1, gradient accumulation 8, bf16 precision, cutoff length 1024, and 3 training
epochs.

For direct unlearning baselines, we use AdamW, learning rate
$5\times10^{-5}$, batch size 2, one epoch, gradient clipping at 1.0, and
bfloat16 model loading. Unless otherwise specified, we set
$\gamma=1.0$, $\alpha=1.0$, and use NLL retain loss. For NPO, we use
$\beta_{\mathrm{NPO}}=1.0$ in the direct baseline. For RMU, the default
steering coefficient is 20.0.

For meta-unlearning, we use inner-loop step size $1\times10^{-5}$, meta step
size $1\times10^{-5}$, $K=4$ inner steps, one inner epoch, five outer epochs,
and gradient clipping at 1.0. The default outer weights are
$\lambda_f=\lambda_r=1.0$. 

\subsection{Compute}
\label{app:compute}

Experiments are run on NVIDIA A100-SXM4-80GB GPUs. Most training and evaluation
commands use one GPU per run. The implementation uses bf16 when supported;
otherwise it falls back to fp16. For meta-unlearning, the optimizer is
8-bit AdamW when available and AdamW otherwise. We do not report a full
project-level compute estimate because total runtime depends on the selected
model, dataset split, and number of evaluated checkpoints.

\section{Limitations}
\begin{wrapfigure}{r}{0.5\columnwidth}
\vspace{-8mm}
\begin{minipage}{0.48\columnwidth}
\begin{algorithm}[H]
\caption{\small Meta-Test Unlearning}
\label{alg:meta_test}
\footnotesize
\begin{algorithmic}[1]

\Require Meta-learned initialization $\theta_T^\star$
\Require Frozen VLM components $\phi$
\Require Meta-test deletion target $z_{\mathrm{test}}$
\Require Text support forget set
$D_{f,\mathrm{sup}}^{\mathrm{text},z_{\mathrm{test}}}$
\Require Optional text retain set $D_{r,\mathrm{sup}}^{\mathrm{text}}$
\Require Meta-test unlearning operator $m_{\mathrm{test}}$
\Require Inner-loop steps $K_{\mathrm{test}}$ and step size $\alpha$

\State Set $\theta_{T,0}^{\mathrm{test}} \leftarrow \theta_T^\star$

\For{$k=0$ to $K_{\mathrm{test}}-1$}
    \State Compute language-backbone unlearning loss
    \[
        \mathcal{L}^{m_{\mathrm{test}}}_{k}
        =
        \mathcal{L}^{m_{\mathrm{test}}}_{\mathrm{in}}
        \left(
            f_{\theta_{T,k}^{\mathrm{test}}};
            D_{f,\mathrm{sup}}^{\mathrm{text},z_{\mathrm{test}}},
            D_{r,\mathrm{sup}}^{\mathrm{text}}
        \right)
    \]

    \State Update language backbone
    \[
        \theta_{T,k+1}^{\mathrm{test}}
        =
        \theta_{T,k}^{\mathrm{test}}
        -
        \alpha
        \nabla_{\theta_{T,k}^{\mathrm{test}}}
        \mathcal{L}^{m_{\mathrm{test}}}_{k}
    \]
\EndFor

\State Recompose the unlearned backbone with frozen VLM:
\[
    F_{\theta_{T,K_{\mathrm{test}}}^{\mathrm{test}},\phi}
\]

\State Evaluate on held-out text-only and multimodal test sets.

\end{algorithmic}
\end{algorithm}
\end{minipage}
\vspace{-12mm}
\end{wrapfigure}
This work has several limitations. First, SMU meta-learns a backbone
initialization through unrolled inner-loop unlearning, which is more expensive
than applying a single unlearning method once. Although test-time unlearning is
cheap, meta-training still requires additional computation. Second, our method
keeps the non-language VLM components frozen. This improves stability, but may
limit forgetting when the target information is mainly stored in the visual
encoder or projector. Third, SMU is evaluated on a limited set of VLMs, datasets,
and unlearning operators. Broader evaluation is needed to confirm whether the
same trends hold for larger models, more diverse visual domains, and more complex
deletion requests. Finally, our results focus on empirical forgetting and utility
metrics. They do not provide a formal guarantee that the target knowledge cannot
be recovered by stronger adversarial prompts or unseen multimodal contexts.

\section{Broader Impact}

This work aims to improve machine unlearning for vision-language models, which
can support privacy protection, deletion requests, and safer model correction.
However, unlearning methods can also be misused to hide undesirable model
behavior or selectively remove accountability-relevant information. In addition,
our method provides empirical forgetting but no formal guarantee that target
knowledge cannot be recovered by stronger adversarial prompts or unseen
multimodal contexts. These risks suggest that VLM unlearning should be used with
careful auditing and transparent evaluation.

\subsection{Assets and Licenses}
\label{app:assets}

We use publicly available datasets, model checkpoints, and baseline methods.
The datasets are Hateful Memes and Covid Memes, and the VLM backbones are
Qwen2.5-VL-7B-Instruct and InternVL3-8B-HF. We cite the original papers or
model releases for all external assets used in the experiments. We do not
release new datasets, pretrained models, or user data.



\newpage

\end{document}